%% file: AOCNNAuthor.tex
\documentclass[acmtog, authorversion, screen=true]{acmartcut} 
\usepackage{mysiggraph2}

\acmSubmissionID{275}

\begin{document}
\title{ Adaptive {O-CNN}: A Patch-based Deep Representation of 3D Shapes}
\author{Peng-Shuai Wang}
\author{Chun-Yu Sun}
\affiliation{%
  \institution{Tsinghua University}
}
\affiliation{%
  \institution{Microsoft Research Asia}
}
\author{Yang Liu}
\author{Xin Tong}
\affiliation{%
  \institution{Microsoft Research Asia}
}

\authorsaddresses{}

\begin{teaserfigure}
  \centering
  \includegraphics[width=0.95\linewidth]{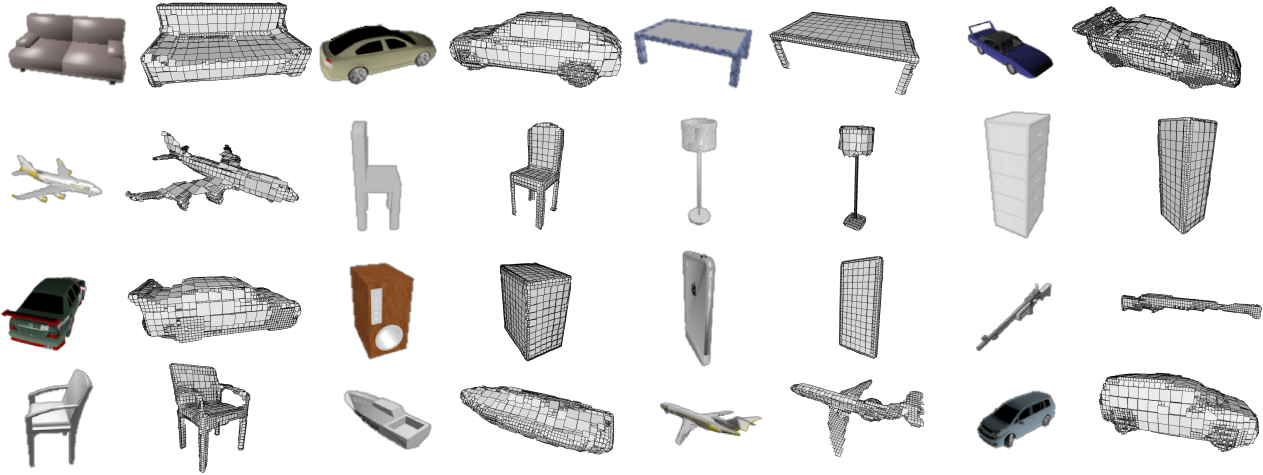}
  \caption{Our Adaptive O-CNN is capable of generating high-quality planar-patch-based shapes from a single image as shown above \rev{(odd columns: input images; even columns: generated shapes)}. The size of the generated planar patches is also adaptively changed according to the underlying predicted geometry. }
  \label{fig:teaser}
\end{teaserfigure}

\begin{abstract}
  We present an Adaptive Octree-based Convolutional Neural Network (Adaptive O-CNN) for efficient 3D shape encoding and decoding. Different from volumetric-based or octree-based CNN methods that represent a 3D shape with voxels in the same resolution, our method represents a 3D shape adaptively with octants at different levels and models the 3D shape within each octant with a planar patch. Based on this adaptive patch-based representation, we propose an Adaptive O-CNN encoder and decoder for encoding and decoding 3D shapes. The Adaptive O-CNN encoder takes the planar patch normal and displacement as input and performs 3D convolutions only at the octants at each level, while the Adaptive O-CNN decoder infers the shape occupancy and subdivision status of octants at each level and estimates the best plane normal and displacement for each leaf octant. \rev{As a general framework for 3D shape analysis and generation, the Adaptive O-CNN not only reduces the memory and computational cost, but also offers better shape generation capability than the existing 3D-CNN approaches}.
  We validate Adaptive O-CNN in terms of efficiency and effectiveness on different shape analysis and generation tasks, including shape classification, \rev{3D autoencoding, shape prediction from a single image}, and shape completion for noisy and incomplete point clouds.
\end{abstract}

\ccsdesc[500]{Computing methodologies~Mesh models}
\ccsdesc[500]{Computing methodologies~Neural networks}

\keywords{Patch-guided adaptive octree, Adaptive O-CNN, shape reconstruction, 3D encoder and decoder }

\maketitle

\input{src2/introduction}

\input{src2/relatedwork}

\input{src2/datastructure}
\input{src2/method}

\input{src2/results}
\input{src2/conclusion}

\begin{acks}
  We wish to thank the authors of ModelNet and ShapeNet for sharing data, Stephen Lin for proofreading the paper, and the anonymous reviewers for their valuable feedback. \looseness=-1
\end{acks}
\bibliographystyle{ACM-Reference-Format}
\bibliography{src2/reference}

\newpage
\input{src2/appendix}


\end{document}

%% file: src2/introduction.tex
\section{Introduction} \label{sec:intro}

3D shape analysis and generation are two key tasks in computer graphics. Traditional approaches generally have limited ability to handle complex shapes, or require significant time and effort from users to achieve acceptable results. With the rapid growth of created and captured 3D data, it has become possible to learn the shape space from a large 3D dataset with the aid of machine learning techniques and guide the shape analysis and generation with the learned features. Recently, deep learning with convolution neural networks has been applied to 3D shape analysis and synthesis. 

Different from images whose grid-based representation is simple and regular, 3D shapes have a variety of representations because of different demands from real applications. For the learning-based shape generation task, the representation of 3D shapes plays a vital role which affects the design of learning architectures and the quality of generated shapes. The commonly used (dense)-\emph{voxel} representation is most popular in existing 3D learning and generation frameworks~\cite{Wu2015}~\cite{Wu2016} since it is a natural extension to 2D images and is well-suited to existing learning frameworks, like convolutional neural networks. However, its high memory storage and costly computation are a major drawback, and high-resolution outputs are hard to produce in practice.
\rev{\emph{Multi-view images}}~\cite{Su2015} have been widely used in shape generation. The generated multi-view images can be fused to reconstruct the complete shape. Proper view selection, enforcing consistency of different views, and shape occlusion are the main challenges for this representation. Recently, \emph{points}, as another common 3D representation, has become a suitable representation for shape analysis and generation with the development of PointNet~\cite{Qi2016} and its variants. However, its output quality is limited by the number of points, and extracting high-quality surfaces from the point cloud requires additional processing. As the favorite 3D format in computer graphics, the \emph{polygonal mesh} has recently been used in learning-based shape generation. Surface patches or meshes can be predicted directly by a neural network that deforms a template mesh or finds a 2D-to-3D mapping~\cite{Groueix2018,kato2018renderer,Wang2018}. However, the predefined mesh topology and the regular tessellation of the template mesh prevent generating high-quality results, especially for irregular and complex shapes. \looseness=-1

The \emph{octree}, which is the most representative sparse-voxel representation, has been integrated with convolution neural networks recently for shape analysis ~\cite{Riegler2017,Wang2017} and its memory and computational efficiency property is suitable for generating high-resolution shapes ~\cite{Tatarchenko2017,Hane2017}. The octree-based generation network usually predicts the occupancy probability of an octant: \emph{occupied}, \emph{free} and \emph{boundary}, and splits the octant with label \emph{boundary}. The prediction and splitting procedures are recursively performed until the predefined max depth of the octree is reached. At the finest level the non-empty leaf octants represent the predicted surface. In existing work, the non-empty leaf octants at the finest level can be regarded as uniform samples of the shape in the x, y, and z directions. We observe that it is actually not necessary to store the shape information in this uniform way since the local shape inside some octants can be represented by simple patches, like planar patches. Therefore, by storing the patch information and terminating the octant split early if the patch associated with the octant well approximates the local shape, the generated octree can have a more compact and adaptive representation. Furthermore, the stored patch has a higher order approximation accuracy than using the center or one of the corners of the octant as the sample of the surface.

Based on the above observations, we propose a novel 3D convolutional neural network for 3D shape called Adaptive Octree-based CNN, or \emph{Adaptive O-CNN} for short. Adaptive O-CNN is based on a novel patch-guided adaptive octree shape representation which adaptively splits the octant according to the approximation error of the estimated simple patch to the local shape contained by the octant. The decoder of Adaptive O-CNN predicts the occupancy probability of octants: \emph{empty}, \emph{surface-well-approximated}, and \emph{surface-poorly-approximated}; infers the local patch at each non-empty octant at each level, and split octants whose label is \emph{surface-poorly-approximated}. It results in an adaptive octree whose estimated local patches at non-empty leaf octants are a multi-scale and adaptive representation of the predicted shape. Besides the decoder, we also develop an efficient 3D encoder for adaptive octrees and use it for shape classification and as a 3D autoencoder. 

Our Adaptive O-CNN inherits the advantages of octree-based CNNs and gains substantial efficiency in memory and computation cost compared with the existing octree-based CNNs due to the use of the adaptive octree data structure. The local patch estimation at each level also enhances the generated shape quality significantly. With all of these features, Adaptive O-CNN is capable of generating high-resolution and high-quality shapes efficiently. We evaluate Adaptive O-CNN on different tasks, including shape classification, 3D autoencoding, shape prediction from a single image, and shape completion for incomplete data. We demonstrate the superiority of Adaptive O-CNN over the state-of-the-art learning-based shape generation techniques in terms of shape quality.

%% file: src2/relatedwork.tex
\section{Related Work} \label{sec:related}

\paragraph{Shape representations for 3D CNNs}
Due to the variety of 3D shape representations, there is not a universal representation for 3D learning. \emph{(dense)-voxel representation} equipped with binary occupancy signals or signed distance values is popular in existing 3D CNN frameworks~\cite{Wu2015,Maturana2015} due to its simplicity and similarity to its 2D counterpart --- images. Voxel-based 3D CNNs often suffer from the high-memory issue, thus they have difficulty in supporting high-resolution input. Since the 3D shape only occupies a small region in its bounding volume, there is a trend toward building \emph{a sparse-voxel representation} for 3D CNNs. A series of works including~\cite{Graham2015, Uhrig2017, Wang2017, Riegler2017} explore the sparsity of voxels and define proper convolution and pooling operations on sparse voxels with the aid of the octree structure and its variants. Our patch-guided adaptive octree also belongs to this type of representation, but with greater sparsity and better accuracy because of its adaptiveness and patch fitting.
\emph{The multi-view representation} regards the shape as a collection of images rendered from different views~\cite{Su2015}. The images can contain RGB color information or view-dependent depth values, and it is easy to feed them to 2D CNNs and utilize networks pretrained on ImageNet~\cite{Deng2009}. However, the multi-view representation may miss partial information of the shape due to occlusion and insufficient views. Recently, \emph{the point-based representation} has become popular due to its simplicity and flexibility. PointNet~\cite{Qi2016} and its successor PointNet++~\cite{qi2017pointnetplusplus} regard a shape as an unorganized point cloud and use symmetric functions to achieve the permutation invariance of points. These point-based CNNs are suited to applications whose input can be well approximated by a set of points or naturally has a point representation, like LiDAR scans. For mesh inputs where the neighbor region is well-defined, \emph{graph-based CNNs}~\cite{Bronstein2016} and \emph{manifold-based CNNs}~\cite{Boscaini2015,Masci2015} find their unique advantages for shape analysis, especially on solving the shape corresponding problem.

\paragraph{3D decoders}
Developing effective 3D decoders is the key to the learning-based shape generation task. The existing work can be categorized according to their shape representations.
\begin{itemize}[leftmargin=*, itemsep=1mm]
  \item [--] \emph{Dense voxel-based decoder}. Brock \etal~\shortcite{Brock2016} proposed a voxel-based variational autoencoder~\cite{Kingma2014} to reconstruct 3D shapes and utilized the trained latent code for shape classification. Choy \etal~\shortcite{Choy2016} combined the power of the 3D volumetric autoencoder and the long short-term memory (LSTM) technique to reconstruct a volumetric grid from single-view or multi-view images. Generative adversarial networks (GAN)~\cite{Goodfellow2016} were introduced to voxel-based shape generation and reconstruction~\cite{Wu2016,Yang18} with different improvement strategies. However, the low resolution of the voxel representation still exists.
  \item [--] \emph{Sparse voxel-based decoder}. The works of~\cite{Tatarchenko2017,Hane2017} show that the octree-based representation offers better efficiency and higher resolution than the (dense)-voxel representation for shape prediction. Riegler \etal~\shortcite{Riegler2017a} demonstrated the usage of the octree-based decoder on depth fusion. $128^3$, $256^3$ and even higher resolution outputs are made possible by octree-based decoders. Our Adaptive O-CNN further improves the efficiency and the prediction accuracy of the octree-based generation network.
  \item [--] \emph{Multi-view decoder}. \rev{Soltani \etal~\shortcite{Soltani2017} proposed to learn a generative model over multi-view depth maps or their corresponding silhouettes, and reconstruct 3D shapes via a deterministic rendering function.} Lun \etal~\shortcite{Lun2017} used an autoencoder structure with a GAN to infer view-dependent depths of a category-specified shape from a single or two sketch inputs and fused all the outputs to reconstruct the shape. Lin \etal~\shortcite{Lin2018} used the projection loss between the point cloud assembled from different views and the ground-truth to further refine the predicted shape.
  \item [--]\emph{ Point-based decoder}. Su \etal~\shortcite{Su2017} designed PointSetGen to predict point coordinates from a single image. The Chamfer distance and Earth Mover's distance metrics are used as the loss functions to penalize the deviation between the prediction and the ground truth. The generated point set roughly approximates the expected shape. Recently, Achlioptas \etal~\shortcite{Achlioptas2018} adapted the GAN technique to improve the point-set generation.
  \item [--] \emph{Mesh-based decoder}. By assuming that the topology of the generated shape is genus-zero or of a disk topology, a series of works~\cite{Sinha2017,Wang2018,kato2018renderer,Yang2018} predicts the deformation of template mesh/point cloud vertices via CNNs. Groueix \etal~\shortcite{Groueix2018} relaxed the topology constraints by introducing multiple 2D patches and predicting the mappings from 2D to 3D. They achieved better quality shapes. However, the uncontrolled distortion by the deformation or the mapping often yields highly irregular and distorted mesh elements that degrade the predicted shape quality.
  \item [--] \emph{Primitive decoder}. Many shapes like human-made objects consist of simple parts. So instead of predicting low-level elements like points and voxels, predicting middle-level or even high-level primitives is essential to understanding the shape structure. \rev{Li \etal~\shortcite{Li2017} proposed a recursive neural network based on an autoencoder to generate the hierarchical structure of shapes. Tulsiani \etal ~\shortcite{ Tulsiani2017} abstracted the input volume by a set of simple primitives, like cuboids, via an unsupervised learning approach. Zou \etal~\shortcite{PRNN2017} built a training dataset where the shapes are approximated by a set of primitives as the ground-truth, and they proposed a generative recurrent neural network to generate a set of simple primitives from a single image to reconstruct the shape.} Sharma \etal~\shortcite{Sharma2018} attempted to solve a more challenging problem: decoding a shape to a CSG tree.
        \rev{We regard a 3D shape as a collection of simple surface patches and use an adaptive octree to organize them for efficient processing. In our Adaptive O-CNN, a simple primitive patch --- planar patch --- is estimated at each octant to approximate the local shape.} 
\end{itemize}

%

\paragraph{Octree techniques}
The octree technique~\cite{Meagher1982} partitions a three-dimensional space recursively by subdividing it into eight octants. It serves as a central technique for many computer graphics applications, like rendering, shape reconstruction and collision detection. Due to its spatial efficiency and friendliness to GPU implementation, the octree and its variants have been used as the shape representation for 3D learning as described above. The commonly-used octant partitioning depends on the existence of the shape inside the octant and the partitioning is performed until the max tree depth is reached, and it usually results in a uniform sampling. Since the shape signal is actually distributed unevenly in the space, an adaptive sampling strategy can be integrated with the octree to further reduce the size of the octree. Frisken \etal~\shortcite{Frisken2000} proposed the octree-based adaptive distance field (ADF) to maintain high sampling rates in regions where the distance field contains fine detail and low sampling rates where the field varies smoothly. They subdivide a cell in which the distance field can not be well approximated by bilinear interpolation of the corner values. The ADF greatly reduces the memory cost and accelerates many processing operations. Our \rev{patch-guided} adaptive octree follows this adaptive principle and uses the fitting error of the local patch to guide the partitioning. The shape is approximated by all the patches at the leaf octants of the octree with a guaranteed approximation accuracy. \looseness=-1

%% file: src2/datastructure.tex
\begin{figure*}[t]
  \centerline{
    \includegraphics[width=0.18\linewidth]{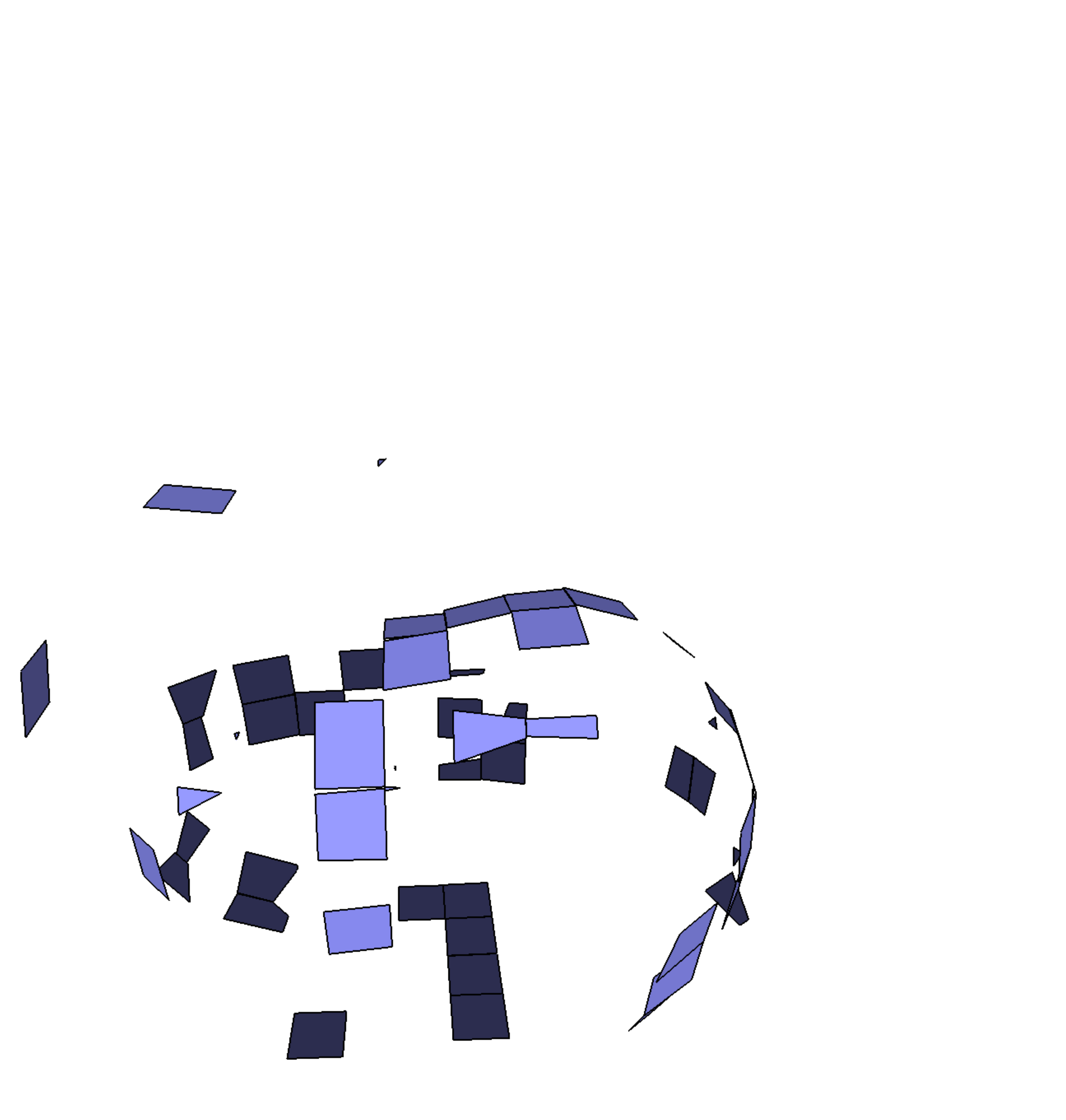} \hfill
    \includegraphics[width=0.18\linewidth]{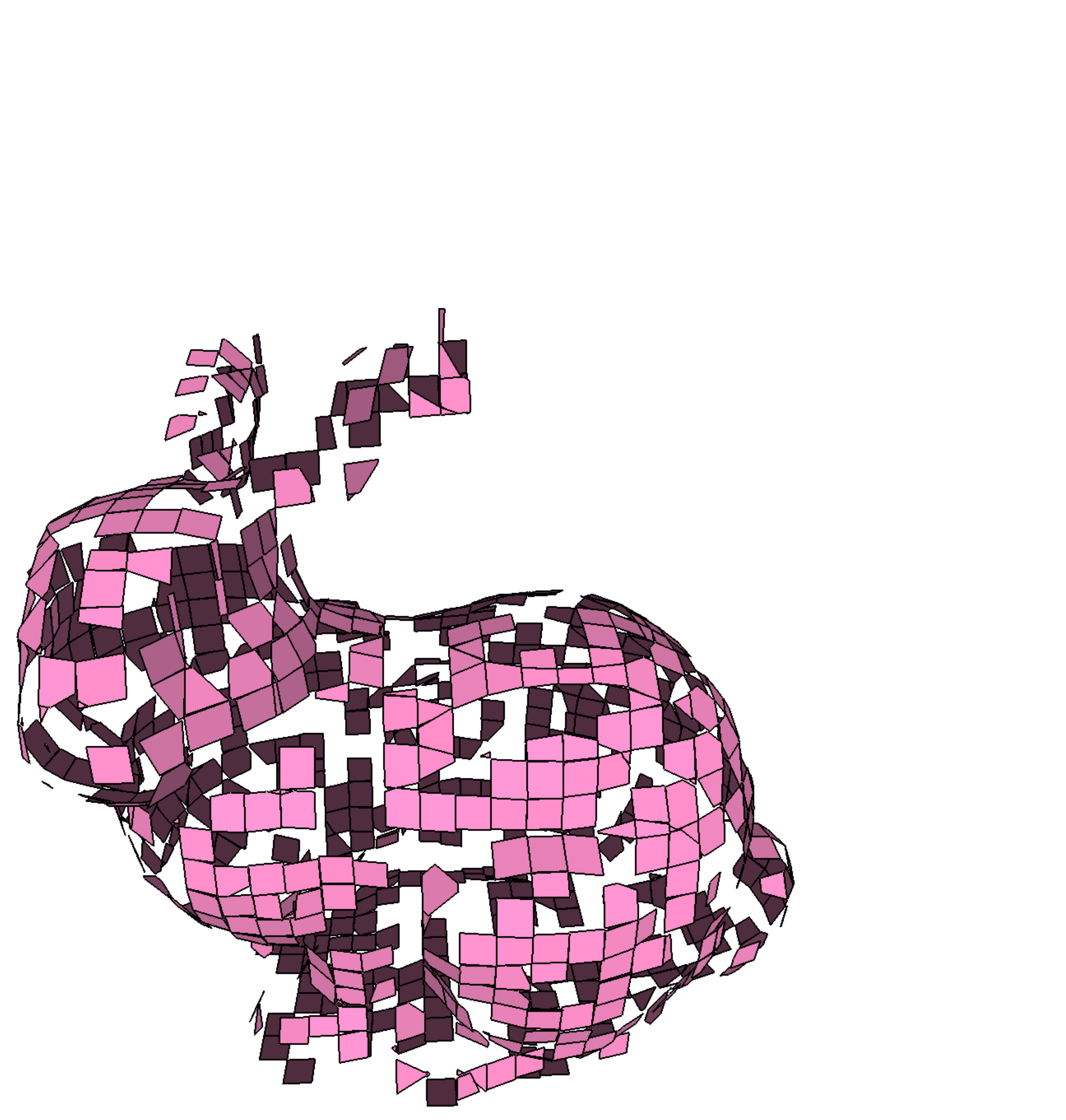}\hfill
    \includegraphics[width=0.18\linewidth]{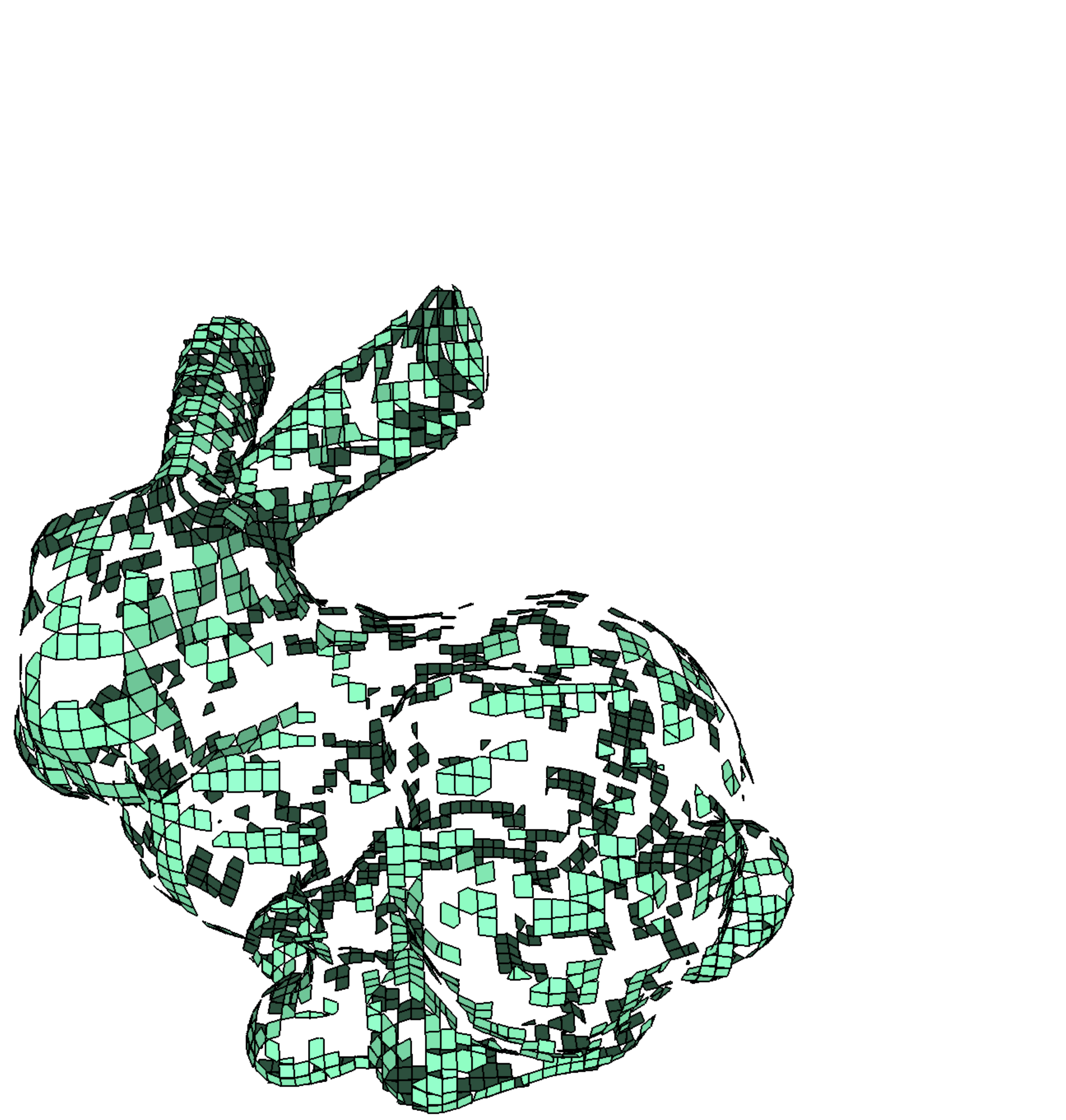}\hfill
    \includegraphics[width=0.18\linewidth]{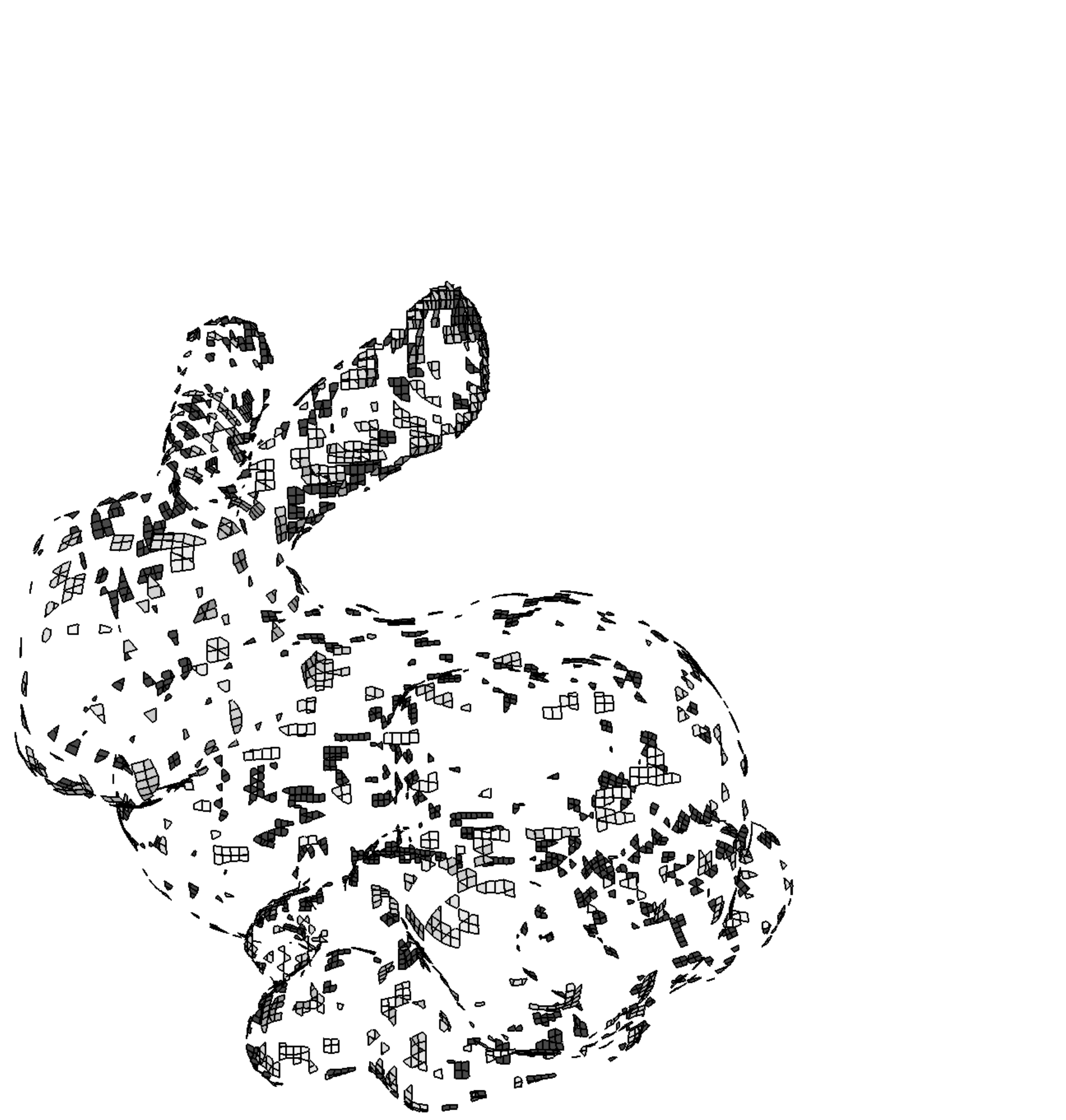}\hfill
    \includegraphics[width=0.18\linewidth]{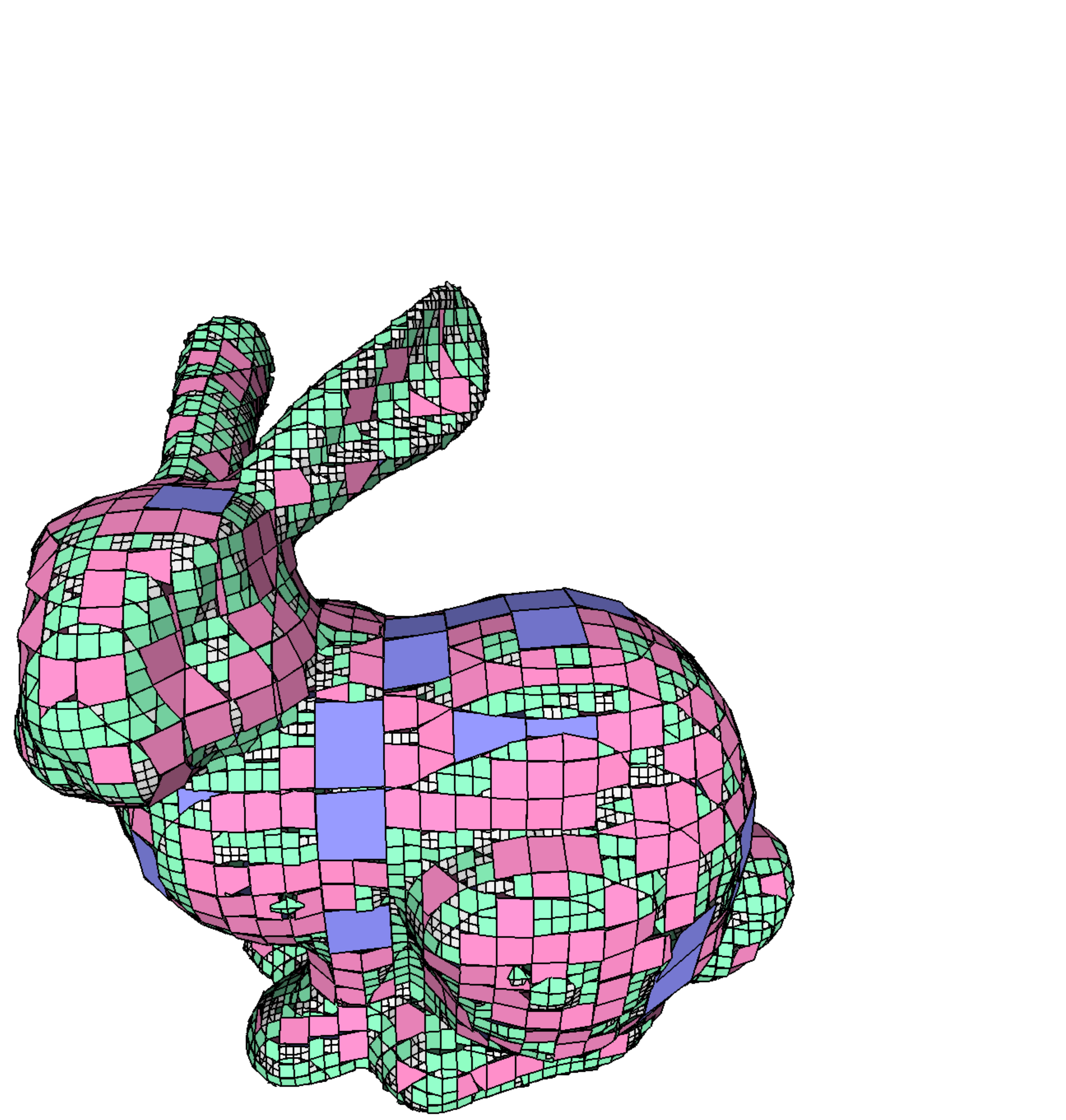}
  }
  \centerline{\small \hspace{0.05\linewidth} (a) 4th-level \hspace{0.13\linewidth} (b) 5th-level \hspace{0.12\linewidth} (c) 6th-level \hspace{0.12\linewidth} (d) 7th-level \hspace{0.08\linewidth} (e) all non-empty leaf nodes}
  \caption{An adaptive octree with planar patches built upon the Bunny model. The max depth of the octree is 7. From left to right: planar patches on the non-empty leaf nodes at the $4$th-level, $5$th-level, $6$th-level, $7$th-level and all planar patches at all the non-empty leaf nodes. The colors on the facets encode the depth level of the octants where planar patches lie. }\label{fig:adpativeoctree} 
\end{figure*}

\begin{figure}[t]
  \centerline{
    \includegraphics[width=1\linewidth]{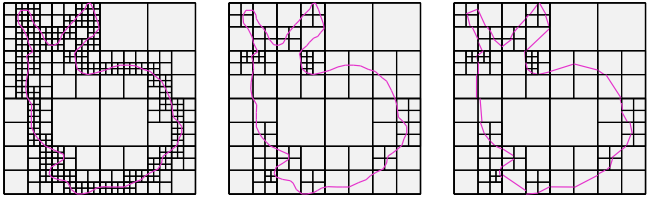}
  }
  \caption{\rev{A simple 2D illustration of the quadtree (left) and the line-segment-guided adaptive quadtree (middle) of a 2D curve. The line segments are illustrated in the rightmost quadtree.}}\label{fig:quadtree}
  \vspace{-6mm}
\end{figure}

\section{Patch-guided Adaptive Octree} \label{sec:datastructure}

We introduce a patch-guided partitioning strategy to generate adaptive octrees. For a given surface $\mS$, we start with its bounding box and perform 1-to-8 subdivision. For octant $\mO$, denote $\mS_\mO$ as the local surface of $\mS$ restricted by the cubical region of $\mO$. If $\mS_\mO \neq \emptyset$, we approximate a \rev{simple surface} patch to $\mS_\mO$. \rev{In this paper, we choose the simplest surface --- a planar patch --- to guide the adaptive octree construction. 
  The best plane $\mP$ with the least approximation error to $\mS_\mO$ is the minimizer of the following objective:
  \begin{equation}\label{eq:plane}
    \Min_{\substack{\mn \in \mathbb{R}^3, d \in \mathbb{R}\\ \|\mn\|=1}} \int_{\mp \in \mS_\mO} \|\mn \cdot \mp + d \|^2 \, \mathrm{d} \mp.
  \end{equation}
  Here \rev{$\mn \in \mathbb{R}^3$} is the unit normal vector of the plane and the plane equation is $ \mP: \mn \cdot \mx + d = 0, \mx \in \mathbb{R}^3$.
  To make the normal direction consistent to the underlying shape normal, we check whether the angle between $\mn$ and the average normals of $\mS_\mO$ is less than $90$ degrees: if not, $\mn$ and $d$ are multiplied by $-1$. In the rest of the paper, we always assume that the local planes are reoriented in this way.
}

We denote $\mP_\mO$ as the planar patch of $\mP$ restricted by the cubical region of $\mO$. The shape approximation quality of the local patch, $\delta_\mO$, is defined by the Hausdorff distance between $\mP_\mO$ and $\mS_\mO$:
$$
  \delta_\mO = \dis\nolimits_{H} (\mP_\mO, \mS_\mO).
$$

The revised partitioning rule of the octree is:
\emph{For any octant $\mO$ which is not at the max depth level, subdivide it if $\mS_\mO \neq \emptyset$ and $\delta_\mO$ is larger than the predefined threshold $\hat{\delta}$.}

By following this rule, a patch-guided adaptive octree can be generated. The patches at all the non-empty leaf octants provide a good approximation to the input 3D shape --- the Hausdorff distance between them and the input is bounded by $\hat{\delta}$. In practice, we set $\hat{\delta} = \frac{\sqrt{3}}{2} h$, where $h$ is the edge length of the finest grid of the octree.


\Cref{fig:adpativeoctree} shows a planar-patch-guided adaptive octree for the 3D Bunny model. We can see that the planar patches are of different sizes due to the adaptiveness of the octree.

For better visualization, we also illustrate the adaptive octree idea in 2D (see \Cref{fig:quadtree}) for a 2D curve input. It is clear that the line-segment-guided adaptive quadtree takes much less memory compared to the quadtree, and the collection of line segments is a good approximation to the input.

\paragraph{Watertight mesh conversion} Due to the approximation error, the local patches between adjacent non-empty leaf octants are not seamlessly connected, \ie gaps exist on the boundary region of octants. This artifact can be found in \Cref{fig:adpativeoctree}(e) \rev{and \Cref{fig:quadtree}-right}. To fill these gaps, surface reconstruction~\cite{Kazhdan2013,Fuhrmann2014} and polygonal repairing~\cite{Ju2004,Attene2013} techniques can be employed. 

%% file: src2/method.tex
\begin{figure*}[t]
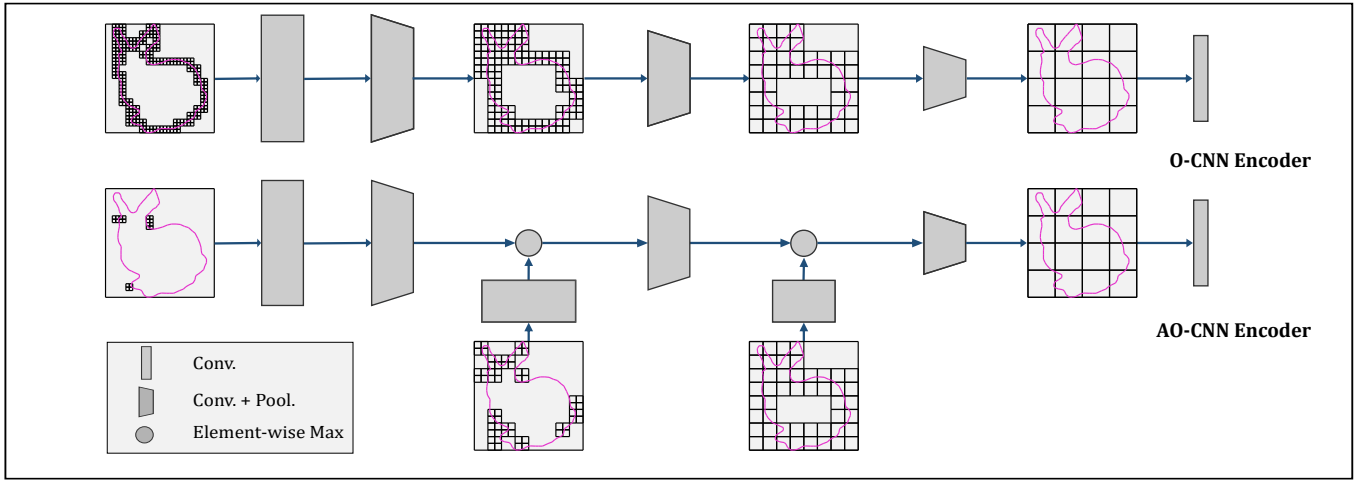

  \centering
  \begin{overpic}[width=1\linewidth]{encoderflow3}
  \end{overpic}
  \caption{The encoder networks of O-CNN and Adaptive O-CNN for an octree with a max-depth 5. Here the 2D adaptive quadtrees are used for illustration purpose only. The ``Conv.'' operation includes ReLU (Rectified Linear Unit) and BN (Batch Normalization). }\label{fig:aocnnencoder} \vspace{-2.2mm}
\end{figure*}

\section{Adaptive O-CNN} \label{sec:aocnn}
The major components of a 3D CNN include the encoder and the decoder, which are essential to shape classification, \rev{shape generation} and other tasks. In \Cref{subsec:encoder} and \Cref{subsec:decoder}, we introduce the 3D encoder and decoder of our adaptive octree-based CNN.\@

\subsection{3D Encoder of Adaptive O-CNN} \label{subsec:encoder}
Since the main difference between the octree and the adaptive octree is the subdivision rule, the efficient GPU implementation of the octree ~\cite{Wang2017} can be adapted to handle the adaptive octree easily. In the following, we first briefly review O-CNN~\cite{Wang2017}, then introduce the Adaptive O-CNN 3D encoder.

\paragraph{Recap of O-CNN encoder} The key idea of O-CNN is to store the sparse surface signal, such as normals, in the finest non-empty octants and constrain the CNN computation within the octree. In each level of the octree, each octant is identified by its shuffled key~\cite{Wilhelms1992}. The shuffled keys are sorted in ascending order and stored in a contiguous array. Given the shuffled key of an octant, we can immediately calculate the shuffled keys of its neighbor octants and retrieve the corresponding neighborhood information, which is essential to implementing efficient CNN operations.
To obtain the parent-children correspondence between the octants in two consecutive octree levels and mark out the empty octants, an additional \emph{Label} array is introduced to record the information for each octant. Common CNN operations defined on the octree, such as convolution and pooling, are similar to volumetric CNN operations. The only difference is that the octree-based CNN operations are constrained within the octree by following the principle:
\emph{``where there is an octant, there is CNN computation''}.
Initially, the shape signal exists in the finest octree level,
then at each level of the octree, the CNN operations are applied sequentially.
When the stride of the CNN operation is 1, the signal is processed with unchanged resolution and it remains in the current octree level.
When the stride of the CNN operation is larger than 1, the signal is condensed and flows along the octree from the bottom to the top. \Cref{fig:aocnnencoder}-upper illustrates the encoder structure of O-CNN.\@

\begin{figure*}[t]
  \centering
  \includegraphics[width=1\linewidth]{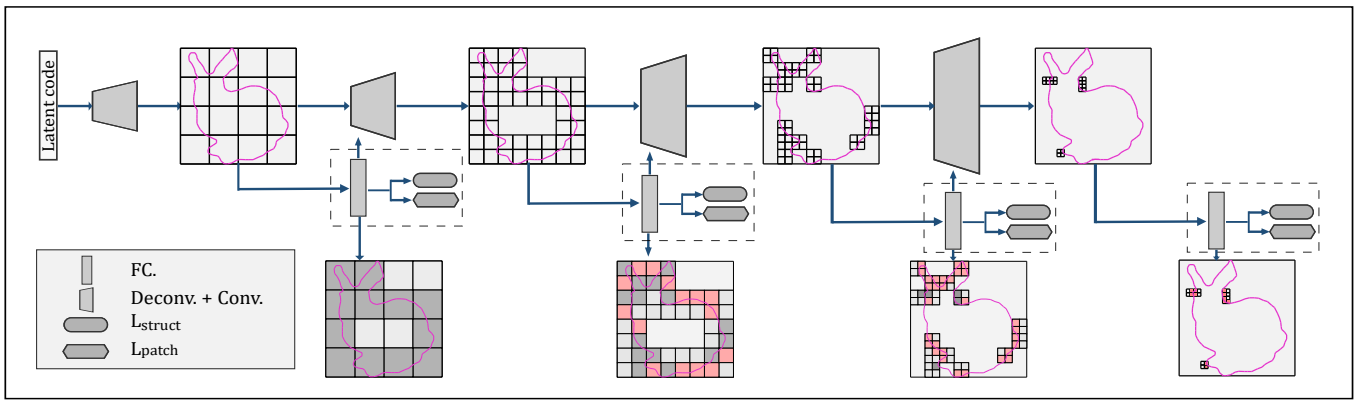}
  \caption{\rev{The decoder network of Adaptive {O-CNN}.
      At each level of the adaptive octree, the \emph{prediction module}, \ie the components inside the dashed boxes, infers the status of the octant:
      \emph{empty}, \emph{surface-well-approximated}, and \emph{surface-poorly-approximated}; and regresses the plane parameters.
      Here we still use 2D adaptive quadtrees for illustration purpose. The three statuses are marked as white, red and gray, respectively. The adaptive planar patches predicted in the red octants represent the generated shape.}
  }\label{fig:paocnndecoder}
  \vspace{-3mm}
\end{figure*}

We reuse the O-CNN's octree implementation for the 3D encoder of Adaptive O-CNN.\@ The data storage of the adaptive octree in the GPU, the convolution, and pooling operations are as same as for O-CNN.\@ There are two major differences between Adaptive O-CNN and O-CNN: (1) the input signal appears at all the octants, not only at the finest octants; (2) the computation starts from leaf octants at different levels simultaneously and the computed features are assembled across different levels. We detail these differences as follows. \looseness=-1

\paragraph{Input signal} Different from O-CNN which only stores the shape signals at the finest octants, we utilize all the estimated local planes as the input signal. For an octant $\mO$ at the $l$-level whose local plane is $\mn \cdot \mx + d = 0, \mx \in \mathbb{R}^3$, we set a four-channel input signal in it: $(\mn, d^\star)$. Here $\mc$ is the center point of $\mO$ and $ d^\star = d - \mn \cdot \mc $. Note that $\mn \cdot(\mx-\mc) + d^\star = 0 $ is \rev{the same plane equation}. Here we use $d^\star$ instead of $d$ because $d^\star$ is bounded by the grid size of $l$-level and it is a relative value while $d$ has a large range \rev{since $d$ measures the distance from the origin to the plane}. For \rev{an} empty octant, its input signal is set to $(0,0,0,0)$.

\paragraph{Adaptive O-CNN 3D encoder} We design a novel network structure to take the adaptive octree as input. On each level of the octree, we apply a series of convolution operators and ReLUs to the features on all the octants at this level and the convolution kernel is shared by these octants. Then the processed features at the $l$-th level is downsampled to the $(l-1)$-th level via pooling and are fused with the features at the $(l-1)$-th level by the \emph{element-wise max} operation. These new features can be further processed and fused with features at the $(l-2)$-th level, $(l-3)$-th level, \ldots, up to the coarsest level. In our implementation, the coarsest level is set to 2, where the octants at the $2$nd-level are enforced to be full so that the features all have the same dimension.
\Cref{fig:aocnnencoder}-lower illustrates our Adaptive O-CNN 3D encoder architecture. \looseness=-1

\subsection{3D Decoder of Adaptive O-CNN}\label{subsec:decoder}
We design a 3D decoder to generate an adaptive octree from a given latent code. The decoder structure is shown in \Cref{fig:paocnndecoder}. At each octree level, we train a neural network to predict \rev{the patch approximation status for each octant} --- \emph{empty}, \emph{surface-well-approximated}, and \emph{surface-poorly-approximated} --- and regress the local patch \rev{parameters}.
Octants with label \emph{surface-poorly-approximated} will be subdivided and their features in them are passed to their children via a \rev{deconvolution operation (also known as ``transposed convolution'' or ``up-convolution'')}.
The label \rev{\emph{surface-well-approximated}} within an octant implies that the local patch can well approximate the local shape where the error is bounded by $\hat{\delta}$, and the network stops splitting at such octants and leaves them as leaf nodes at the current octree level. An adaptive octree can be \rev{predicted in this recursive way until the max depth is reached}.

\paragraph{Prediction module} The neural network for prediction is simple. It consists of ``FC + BN + ReLU + FC'' operations. Here BN represents \emph{Batch Normalization} and FC represents \emph{Fully Connected layer}. This module is shared across all octants at the same level of the adaptive octree. The output of the prediction module includes the patch approximation status and the plane patch parameters $(\mn, d^\star)$. \rev{The patch approximation status guides the subdivision of the octree and is also required by the octree-based deconvolution operator}.


\paragraph{Loss function} The loss function of the Adaptive O-CNN decoder includes the structure loss and the patch loss. The structure loss $L_{\texttt{struct}}$ measures the difference between the predicted octree structure and its ground-truth.
Since the determination of octant status is a 3-class classification, we use the cross entropy loss to define the structure loss.
Denote $H_l$ as the cross entropy at $l$-level of the octree, the structure loss is formed by the weighted sum of cross entropies across all the levels:
\begin{align}
  L_{\texttt{struct}} = \sum_{l=2}^{l_{\max}} \dfrac{w_l}{n_l}H_l.
\end{align}
Here $n_l$ is the number of octants at the $l$-th level of the predicted octree, $l_{\max}$ is the max depth, and $w_l$ is the weight defined on each level. \rev{Similar to the encoder, the coarsest level of the octree is 2 and is full of octants, so $l$ starts with $2$ in the above equation.} In our implementation, we set $w_l$ to 1.

The patch loss $L_{\texttt{patch}}$ measures the squared distance error between the plane parameters and the ground truth at all the leaf octants in each level:
\begin{align}
  L_{\texttt{patch}} = \sum_{l=4}^{l_{\max}} \dfrac{w_l}{n^\prime_l} \sum_{i=1}^{n^\prime_l} \lambda \| \mn_i - \bar{\mn}_i\|^2 + | d_i^\star - \bar{d}_i^\star |^2.
\end{align}
Here $\mn_i$ and $d_i^\star$ are the predicted parameters, $\bar{\mn}_i$ and $\bar{d}_i^\star$ are the corresponding ground-truth values, and $n^\prime_l$ is the number of leaf octants at the $l$-th level of the predicted octree,  $\lambda$ is set to $0.2$. 
In our implementation we make the octree adaptive when the octree level is over 4, so $l$ starts with $4$ in the above equation. 
\rev{Note that for the wrongly generated octants that do not exist in the ground-truth, there is no patch loss for them, and they are penalized by the structure loss only.}

We use $L_{\texttt{struct}} + L_{\texttt{patch}} $ as the loss function for our decoder.
Since the predicted plane should pass through the octant (otherwise it violates the assumption that the planar patch is inside the octant cube), we add the constraint $|d^\star| < \frac{\sqrt{3}}{2} h_l$ where $h_l$ is the grid size of $l$-level octants, by utilizing the $\tanh$ function on the network output.

%% file: src2/results.tex
\section{Experiments and Comparisons} \label{sec:result}

To evaluate Adaptive O-CNN, we conduct three experiments:
3D shape classification, 3D autoencoding and 3D shape prediction from a single image.
All the experiments were done on a desktop computer with an Intel Core I7-6900K CPU (3.2GHz)
and a GeForce GTX Titan X GPU (12 GB memory).
Our implementation is based on the Caffe framework~\cite{Jia2014} \rev{and
  the source code is available at \url{https://github.com/Microsoft/O-CNN}.
  The detailed Adaptive O-CNN network configuration is provided in the supplemental material.} \looseness=-1

\paragraph{Dataset pre-processing} \rev{For building the training dataset for our experiments, we first follow the approach of~\cite{Wang2017} to obtain a dense point cloud with oriented normals by virtual 3D scanning}, then we build the planar-patch-guided adaptive octree from it via the construction procedure introduced in~\Cref{sec:datastructure}.

\subsection{Shape classification}
We evaluate the efficiency and efficacy of our Adaptive O-CNN encoder on the 3D shape classification task.

\paragraph{Dataset}
We performed the shape classification task on the ModelNet40 dataset~\cite{Wu2015}, which contains 12,311 well annotated CAD models from 40 categories.
The training data is augmented by rotating each model along its upright axis  at 12 uniform intervals.
The planar-patch-guided adaptive octrees are generated with different resolutions: $32^3$, $64^3$, $128^3$, and $256^3$. We conducted the shape classification experiment on these data respectively.

\paragraph{Network configuration} To clearly demonstrate the advantages of our adaptive octree-based encoder over the O-CNN~\cite{Wang2017},
we use the same network parameters of O-CNN including the parameters of CNN operations, the number of training parameters and the dropout strategy.
The only difference is the encoder network structure as shown in \Cref{fig:aocnnencoder}.
After training, we use the orientation pooling technique~\cite{Qi2016a,Su2015} to vote for the results from the 12 predictions of the same object under different poses.

\begin{table}[t]
  \centering
  \scalebox{0.8}{
    \begin{tabular}{r|r|rrrrr}
      \toprule
                                & Method         & $32^3$             & $64^3$            & $128^3$           & $256^3$          \\ \midrule
      \multirow{3}{*}{Memory}   & Voxel          & 0.71\,GB           & 3.7\,GB           & ---               & ---              \\
                                & O-CNN          & 0.58\,GB           & 1.1\,GB           & 2.7\,GB           & 6.4\,GB          \\
                                & Adaptive O-CNN & \textbf{ 0.51}\,GB & \textbf{0.95}\,GB & \textbf{1.5}\,GB  & \textbf{1.7}\,GB \\ \midrule
      \multirow{3}{*}{Time}     & Voxel          & 425\,ms            & 1648\,ms          & ---               & ---              \\
                                & O-CNN          & 41\,ms             & 117\,ms           & 334\,ms           & 1393\,ms         \\
                                & Adaptive O-CNN & \textbf{34}\,ms    & \textbf{63}\,ms   & \textbf{ 112}\,ms & \textbf{307}\,ms \\ \midrule
      \multirow{2}{*}{Accuracy} & O-CNN          & 90.4\%             & \textbf{90.6\%}   & \textbf{90.1\% }  & \textbf{90.2\%}  \\
                                & Adaptive O-CNN & \textbf{90.5\%}    & 90.4\%            & 90.0\%            & \textbf{90.2\%}  \\
      \bottomrule
    \end{tabular}}
  \caption{Statistics of the performance of O-CNN and Adaptive O-CNN on the shape classification task. GPU-memory and time consumption on batch size 32, and classification accuracy of each method are reported.
    \rev{Here we remeasured the running time of O-CNN on a Titan X GPU.\@
      For reference, the time and memory cost of the voxel-based 3D CNN are also provided, which was measured on a GeForce 1080 GPU.} 
  }
  \label{table:classification}
\end{table}

\begin{table}[t]
  \centering
  \scalebox{0.67}{
    \begin{tabular}{l|c||l|c}
      \toprule
      Method                        & Accuracy & Method                                   & Accuracy \\
      \midrule
      PointNet~\cite{Qi2016}        & 89.2\%   & PointNet++~\cite{qi2017pointnetplusplus} & 91.9\%   \\
      VRN Ensemble~\cite{Brock2016} & 95.5\%   & SubVolSup~\cite{Qi2016a}                 & 89.2\%   \\
      OctNet~\cite{Riegler2017}     & 86.5\%   & O-CNN~\cite{Wang2017}                    & 90.6\%   \\
      Kd-Network~\cite{Klokov2017}  & 91.8\%   & Adaptive O-CNN                           & 90.5\%   \\
      \bottomrule
    \end{tabular}
  }
  \caption{\rev{ModelNet40 classification benchmark.
      The classification accuracy of Adaptive O-CNN is better and comparable to existing learning-based methods like SubVolSup, PointNet, OctNet and O-CNN, but it is worse than PointNet++, Kd-Network and VRN Ensemble.
    }}
  \label{table:cls}
\end{table}

\begin{table*}[t]
  \centering
  \resizebox{0.85\textwidth}{!}{%
    \begin{tabular}{l|l|lllllllllllll}
      \toprule
                           & mean          & pla.          & ben.          & cab.                & car           & cha.          & mon.          & lam.          & spe.          & fir.          & cou.                & tab.          & cel.                & wat.          \\ \midrule
      PSG                  & 1.91          & 1.11          & 1.46          & 1.91                & 1.59          & 1.90          & 2.20          & 3.59          & 3.07          & 0.94          & 1.83                & 1.83          & 1.71                & 1.69          \\
      AtlasNet(25)         & 1.56          & 0.87          & 1.25          & 1.78                & 1.58          & 1.56          & 1.72          & 2.30          & 2.61          & 0.68          & 1.83                & 1.52          & 1.27                & 1.33          \\
      AtlasNet(125)        & 1.51          & \textbf{0.86} & \textbf{1.15} & 1.76                & 1.56          & \textbf{1.55} & 1.69          & \textbf{2.26} & 2.55          & \textbf{0.59} & 1.69                & 1.47          & 1.31                & \textbf{1.23} \\
      Adaptive O-CNN       & \textbf{1.44} & 1.19          & 1.27          & 1.01                & \textbf{0.96} & 1.65          & \textbf{1.41} & 2.83          & \textbf{1.97} & 1.06          & \textbf{1.14}       & \textbf{1.46} & \textbf{0.73}       & 1.82          \\
      \midrule
      O-CNN(binary)        & 1.60          & 1.12          & 1.30          & 1.06                & 1.02          & 1.79          & 1.62          & 3.71          & 2.56          & 0.98          & 1.17                & 1.67          & 0.79                & 1.88          \\
      O-CNN(patch)         & 1.59          & 1.10          & 1.29          & 1.06                & 1.02          & 1.79          & 1.60          & 3.70          & 2.55          & 0.97          & 1.18                & 1.66          & 0.79                & 1.87          \\
      \rev{O-CNN(patch*) } & \rev{1.53}    & \rev{1.09}    & \rev{1.31}    & \rev{\textbf{0.91}} & \rev{0.97}    & \rev{1.77}    & \rev{1.58}    & \rev{3.64}    & \rev{2.28}    & \rev{0.97}    & \rev{\textbf{1.14}} & \rev{1.65}    & \rev{\textbf{0.73}} & \rev{1.91}    \\ \bottomrule
    \end{tabular}
  }
  \caption{Statistics of 3D Autoencoder experiments. Both the PSG and AtlasNet results are those provided in ~\cite{Groueix2018}. The number after AtlasNet is the number of mesh patches it used. The Chamfer distance is multiplied by $10^3$ for better display. } \vspace{-7mm}
  \label{table:autoencoder}
\end{table*}

\paragraph{Experimental results}
\rev{
  We record the peak memory consumption and the average time of one forward and backward iteration with batch size 32,
  and report them with the classification accuracy on the test dataset in Table~\ref{table:classification}.
  The experiments show that the classification accuracy of Adaptive O-CNN is comparable to O-CNN under all the resolutions,
  and the memory and computational cost of Adaptive O-CNN is significantly lower, especially on the high-resolution input: Adaptive O-CNN under $256^3$ resolution gains about a 4-times speed-up and reduces GPU memory consumption by 73\% compared to O-CNN.
  Compared with state-of-the-art learning-based methods, the classification accuracy of Adaptive O-CNN is also comparable (see Table~\ref{table:cls}).
}

\paragraph{Discussion} As seen from Table~\ref{table:classification},
\rev{when the input resolution is beyond $128^3$, the classification accuracy of Adaptive O-CNN drops slightly.
  We find that when the input resolution increases from $32^3$ to $128^3$,
  the training loss decreases from 0.168 to 0.146, whereas the testing loss increases from 0.372 to 0.375.
  We conclude that Adaptive O-CNN with a deeper octree tends to overfit the training data of ModelNet40.
  The result is also consistent with the observation of \cite{Wang2017} on O-CNN.\@
  With more training data, for instance, by rotating each training object 24 times around their upright axis uniformly,
  the classification accuracy can increase by 0.2\% under the resolution of $128^3$.}

\subsection{3D Autoencoding}

\rev{
  The Autoencoder technique is able to learn a compact representation for the input and recovers the signal from the latent code via a decoder.
  We use the Adaptive O-CNN encoder and decoder presented in \Cref{sec:aocnn} to form a 3D autoencoder.
}

\paragraph{Dataset} We trained our 3D autoencoder on the ShapeNet Core v2 dataset~\cite{shapenet2015},
which consists of 39,715 3D models from 13 categories. The training and the test splitting rule is the same as the ones used in AtlasNet~\cite{Groueix2018} and
the point-based decoder (PSG) ~\cite{Su2017}. \rev{The adaptive octree we used is of max-depth 7, \ie, the voxel resolution is $128^3$.}

\paragraph{Quality metric}
We evaluate the quality of the decoded shape via measuring the Chamfer distance between it and its ground-truth shape.
With the ground-truth point cloud denoted by $S_g = \{x_i\}_{i=1}^n$,
and the points predicted by the neural network by $S = \{\hat{x}_i\}_{i=1}^m$,
the Chamfer distance between $S_g$ and $S$ is defined as:
\[
  D(S_g, S) = \frac{1}{n} \sum_{x_i \in S_g} \min_{\hat{x}_j \in S} \| x_i - \hat{x}_j \|_2^2 +
  \frac{1}{m} \sum_{\hat{x}_i \in S} \min_{x_j \in S_g} \| \hat{x}_i - x_j \|_2^2.
\]
Because our decoder outputs a patch-guided adaptive octree, to calculate the Chamfer distance,
we sample a set of dense points uniformly from the estimated planar patches:
we first subdivide the planar patch contained in the non-empty leaf node of the adaptive octree
towards the resolution of $128^3$, then randomly sample one point on each of the subdivided planar patches to form the output point cloud.
For the ground-truth dense point cloud, we also uniformly sample points from it under the resolution of $128^3$.

\paragraph{Experimental results}
The quality measurement is summarized in Table~\ref{table:autoencoder}, and we also compare with two state-of-the-art 3D autoencoder methods: AtlasNet~\cite{Groueix2018} that generates a set of mesh patches as the approximation of the shape, and PointSetGen (PSG)~\cite{Su2017} that generates a point cloud output.
Compared with two types of AtlasNets which predict 25 and 125 mesh patches, respectively, our Adaptive O-CNN autoencoder achieves the best quality \rev{on average}.
Note that the loss function of AtlasNet is the Chamfer distance exactly, while our autoencoder has not been specified for this loss but still performs well.
\rev{Compared with PSG~\cite{Su2017}, it is clear that our method and AtlasNet are much better.}

\begin{figure*}[t]
  \centering
  \includegraphics[width=0.92\linewidth]{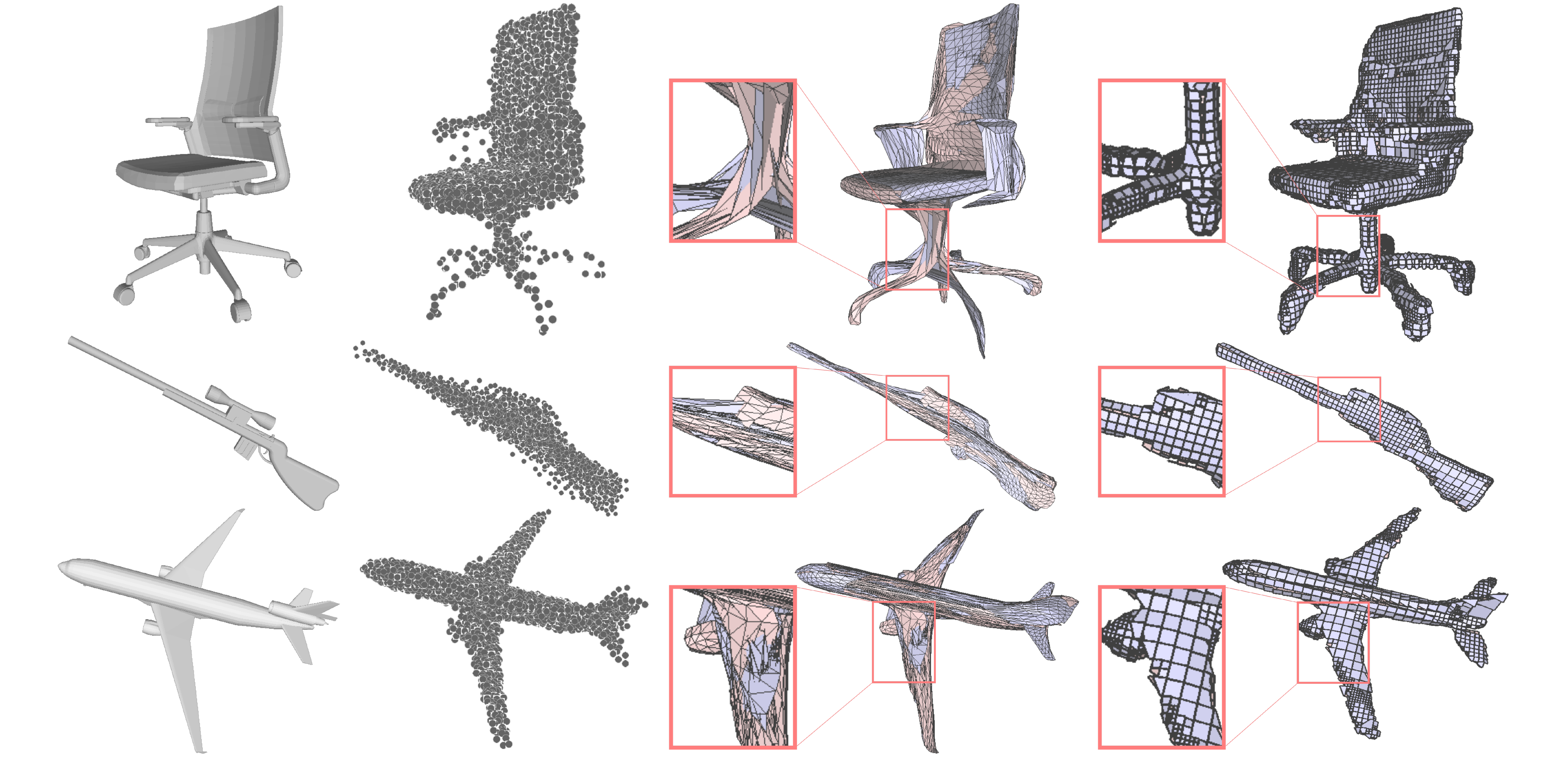}
  \centerline{\small \hfill (a) Input shape \hspace{0.09\linewidth} (b) PSG \hspace{0.18\linewidth} (c) AtlasNet \hspace{0.18\linewidth} (d) Our results \hfill }
  \caption{Visualizations of decoded shapes by different methods. The point clouds are rendered as dots.
    \rev{For the results of AtlasNet and our Adaptive O-CNN, the two sides of the mesh patches are rendered with different colors. It is clear that AtlasNet does not generate orientation-consistent patches. Compared with AtlasNet, the results of our Adaptive O-CNN are much more regular, the normal orientation are also more consistent, and the volume structures are reconstructed better.}
  }
  \label{fig:autoencoder} 
\end{figure*}

\paragraph{Discussion}
\rev{
  Our Adaptive O-CNN performs worse than AtlasNet in some categories, such as plane, chair and firearm.
  We found on relatively thin parts of the models in those categories, such as the wing of the plane, arm of the chair, as well as the barrel of the gun, our Adaptive O-CNN has a larger geometry deviation from the original shape. However, for AtlasNet, although its deviation is smaller, we found that it approximates the thin parts with a single patch or messy patches (e.g. the single patch for the right arm of the chair, the folded patches for the gun barrel and the plane wing as seen in \Cref{fig:autoencoder}), and the volume structure of the thin parts is totally lost and it is difficult or even impossible to define the inside and outside on those regions. We conclude that the Chamfer distance loss function used in AtlasNet does not penalize this structure loss.
  On the contrary, because our Adaptive O-CNN is trained with both the octree-based structure loss and patch loss, it successfully approximates the thin parts with better volume structures (e.g. the cylinder like shape for the gun barrel, chair support and the two-side surfaces for the plane wing). The zoom-in images in \Cref{fig:autoencoder} highlight these differences.}

\begin{figure}[t]
  \centering
  \begin{overpic}[width=1\linewidth]{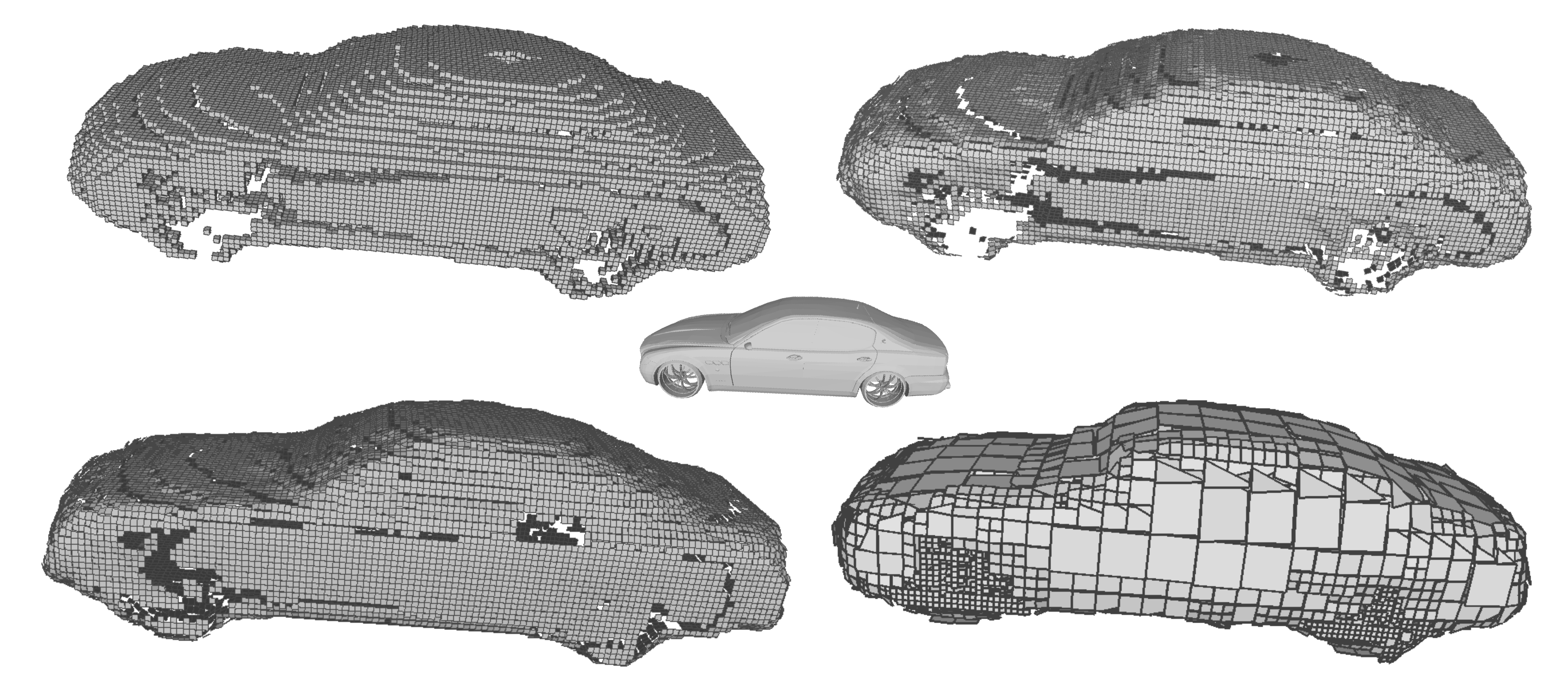}
    \put(12,22){\small (a) O-CNN(binary)}
    \put(63,22){\small (b) O-CNN(patch)}
    \put(12,-3){\small (c) O-CNN(patch*)}
    \put(63,-3){\small (d) Our result}
  \end{overpic}
  \vspace{0.2mm}
  \caption{Ablation study. The shapes from O-CNN(patch) and O-CNN(patch*) are smoother than O-CNN(binary), which shows that the regression of plane patches enables sub-voxel precision. Compared with O-CNN(patch) and O-CNN(patch*), there are less missing regions in the output of Adaptive O-CNN, which verifies the benefit of the adaptive patch based octree representation. The ground-truth shape is shown in the middle.
  }
  \label{fig:ablation} \vspace{-3mm}
\end{figure}
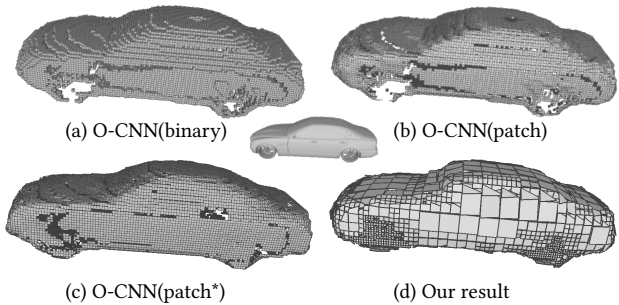

\paragraph{Ablation study}
\rev{
  We designed three baseline autoencoders based on the standard octree structure to demonstrate the need for using the adaptive octree:
  \begin{enumerate}[leftmargin=*]
    \item \emph{O-CNN(binary)}: A vanilla octree based autoencoder. The encoder is presented in \Cref{fig:aocnnencoder}-upper.
          The decoder is similar to the Adaptive O-CNN decoder but with two differences: (1) the prediction module only predicts whether a given octant has an intersection with the ground-truth surface.
          If intersected, the octant will be further subdivided; (2) the loss only involves the structure loss.
    \item \emph{O-CNN(patch)}: An enhanced version of O-CNN(binary). The prediction module also predicts the plane patch on each leaf node at the finest level and the patch loss is added.
    \item \emph{O-CNN(patch*)}: An enhanced version of O-CNN(patch). The prediction module predicts the plane patch on each leaf node at each level and the patch loss is added.
  \end{enumerate}
}

\rev{
  These three networks are trained on the 3D autoencoding task. The statistics of the results are also summarized in Table~\ref{table:autoencoder}.
  The Chamfer distance metric of O-CNN(patch) is slightly better than O-CNN(binary) since the regression of plane patches at the finest level enables sub-voxel precision.
  By considering the patch loss at each depth level, O-CNN(patch*) further improves the reconstruction accuracy due to the hierarchical supervision in the training. However, it is still worse than Adaptive O-CNN. The reason is as follows: during the shape generation of Adaptive O-CNN, if the plane patch generated in an octant in the coarser level can well approximate the ground-truth shape, the Adaptive O-CNN will stop subdividing this octant and the network layers in the finer level are trained to focus on the region with more geometry details. As a result, the Adaptive O-CNN not only avoids the holes in the region that can be well approximated by a large planar patch, but also generates better results for the surface region with more shape details. On the contrary, no matter whether a region can be modeled by a large plane patch or not, the O-CNN based networks subdivide all non-empty octants at each level and predict the surface at the finest level. Therefore, the O-CNN has more chances to predict the occupancy of the finest level voxels wrongly.
}

The visualization in \Cref{fig:ablation} also demonstrates that Adaptive O-CNN generates more visually pleasing results and outputs large planar patches on flat regions, while the outputs of O-CNN(binary), O-CNN(patch) and O-CNN(patch*) contain more holes due to the inaccurate prediction.

\begin{figure}
  \centering
  \includegraphics[width=1\linewidth]{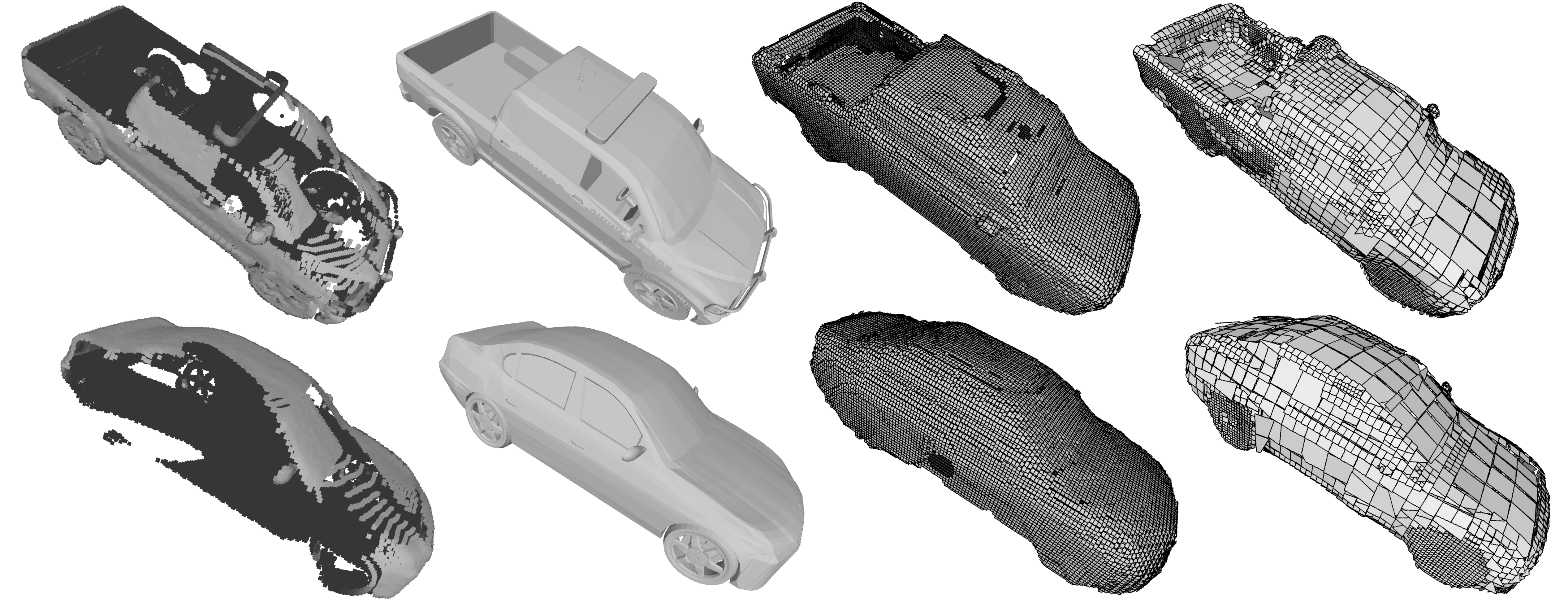}
  \centerline{ \small \hspace{0.01\linewidth} (a) Incomplete shape \hspace{0.005\linewidth} (b) Ground-truth \hspace{0.008\linewidth} (c) O-CNN(patch) \hspace{0.008\linewidth} (d) Our results }
  \caption{Shape completion on two incomplete cars.
    Our completion results are closer to the ground-truth, whereas there are still some missing regions with the O-CNN(patch) based decoder.
    \rev{The Chamfer distance metrics of the two cars are 0.0713 and 0.0349 for O-CNN(patch), 0.0626 and 0.0306 for Adaptive O-CNN.}
  }
  \label{fig:completion} \vspace{-3mm}
\end{figure}

\paragraph{Application: shape completion}
A 3D autoencoder can be used to recover the missing part of a geometric shape and fair the noisy input.
We conduct a shape completion task to demonstrate the efficacy of our Adaptive O-CNN.
We choose the car category from the ShapeNet Core v2 dataset as the ground-truth data.
For each car, we choose 3 to 5 views randomly and sample dense points from these views.
On each view, we also randomly crop some regions to mimic holes and perturb point positions slightly to model scan noise.
These views are assembled together to serve as the incomplete and noisy data.
We trained our Adaptive O-CNN based autoencoder on this synthetic dataset with the incomplete shape as input and the corresponding complete shape as the target.
For reference, we also trained the O-CNN(patch) based autoencoder on it. \rev{The max depth of the octree in all the networks is set to 7.}
Figure~\ref{fig:completion} shows two completion examples.
The results from Adaptive O-CNN are closer to the ground-truth, while the O-CNN(patch) misses filling some holes. \looseness=-1

\subsection{Shape reconstruction from a single image}
Reconstructing 3D shapes from 2D images is an important topic in computer vision and graphics.
With the development of 3D deep learning techniques, the task of inferring a 3D
shape from a single image has gained much attention in the research community.
We conduct experiments on this task for our Adaptive O-CNN and compare it with the state-of-art methods~\cite{Groueix2018,Su2017,Tatarchenko2017}.


\paragraph{Dataset}
For the comparisons with the AtlasNet~\cite{Groueix2018} and PointSetGen (PSG)~\cite{Su2017},
we use the same dataset which is originally from~\cite{Choy2016}.
The ground-truth 3D shapes come from ShapeNet Core v2~\cite{shapenet2015},
and each object is rendered from 24 viewpoints with a transparent background.
For the comparison with OctGen~\cite{Tatarchenko2017}, since OctGen was only trained on the car category with the octree of resolution $128^3$, we also trained our network on the car dataset with the same resolution.

\paragraph{Image encoders}
For the image encoder, AtlasNet~\cite{Groueix2018} used ResNet18~\cite{He2016} and OctGen used the classic LeNet~\cite{Lecun1998}.
We also use ResNet18 and LeNet as the image encoders in the respective comparison for fairness.

\begin{figure*}
  \centering
  \begin{overpic}[width=1\linewidth]{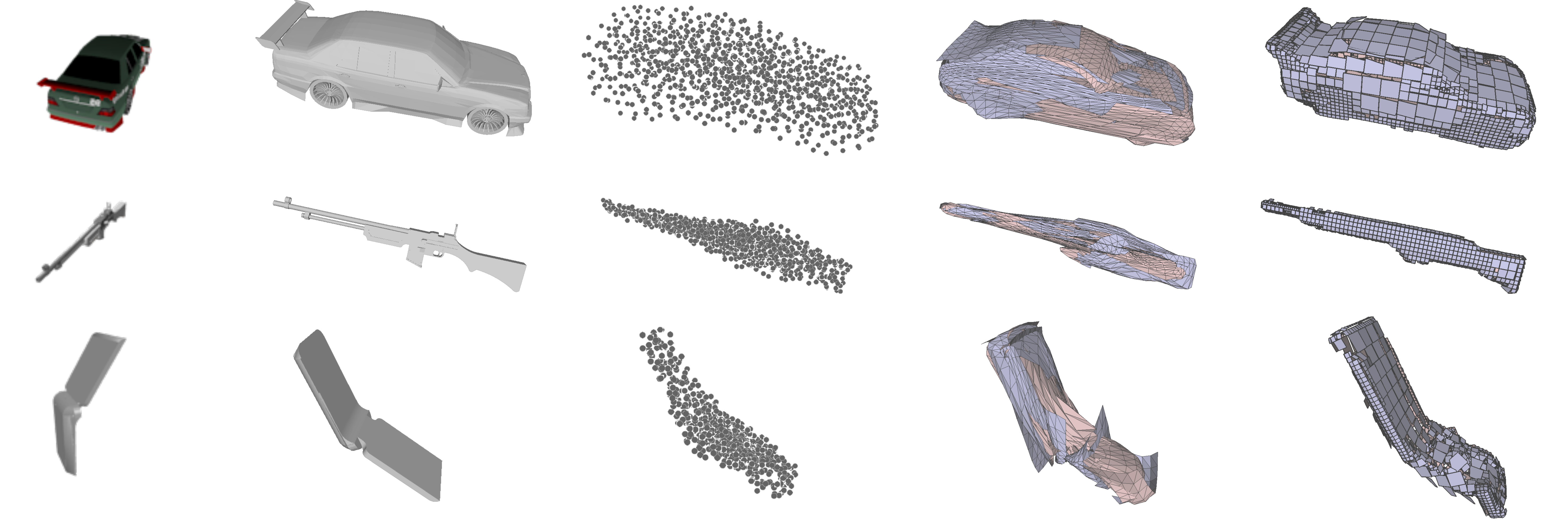}
    \put( 1,-2){\small (a) Input image}
    \put(19,-2){\small (b) Ground-truth}
    \put(43,-2){\small (c) PSG}
    \put(65, -2){\small (d) AtlasNet}
    \put(87,-2){\small (e) Our results}
  \end{overpic}
  \vspace{\baselineskip}
  \caption{Visualizations of shape prediction from a single image. Our Adaptive O-CNN generates more detailed geometry, like the tail wing of the car and the trigger of the rifle.
    \rev{The two sides of the mesh are rendered with different colors, and it is clear that AtlasNet does not generate orientation-consistent patches.}
  }
  \label{fig:image2shape}
\end{figure*}

\paragraph{Experimental results}
We report the Chamfer distance between the predicted points and points sampled from the
original mesh for PointSetGen, AtlasNet and our method in Table~\ref{table:image2shape:chamfer}.
As mentioned in~\cite{Groueix2018}, they randomly selected 260 shapes (20 per category) to form the testing database.
To compare with PointSetGen, they ran the ICP algorithm~\cite{Besl1992}
to align the predicted points from both PointSetGen and AtlasNet with the ground-truth point cloud.
Note that after the ICP alignment, the Chamfer distance error is slightly improved.
To have a fair comparison, we also ran the ICP algorithm to align our results with the ground-truth.
Our method achieves the best performance on 8 out of 13 categories, especially for the objects with large flat regions, such as car and cabinet.
In \Cref{fig:image2shape} \& \Cref{fig:teaser} we illustrate some sample outputs from these networks. It is clear that our outputs are more visually pleasing.
For the flip phone image in the last row of \Cref{fig:image2shape}, the reconstruction quality is relatively lower than other input images for all methods.
This is because flip phones are rare in the training dataset.

\begin{figure*}
  \centering
  \begin{overpic}[width=0.98\linewidth]{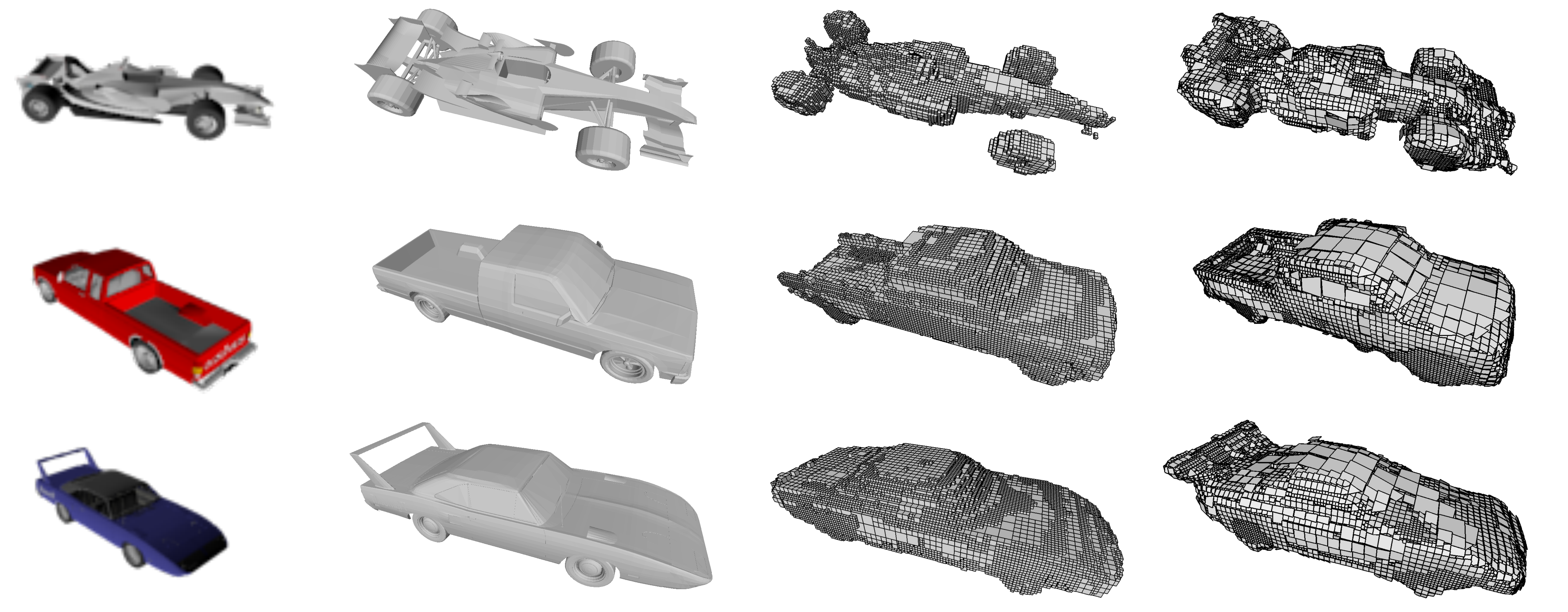}
    \put( 3,-2){\small (a) Input image}
    \put(28,-2){\small (b) Ground-truth}
    \put(56,-2){\small (c) OctGen}
    \put(82,-2){\small (d) Our results}
  \end{overpic}
  \vspace{\baselineskip}
  \caption{Comparison with OctGen on the shape prediction from a single image task. Note that some geometry features are missing in the results of OctGen, like the tail wing of the car (third row).}
  \label{fig:image2shape-ogn} 
\end{figure*}
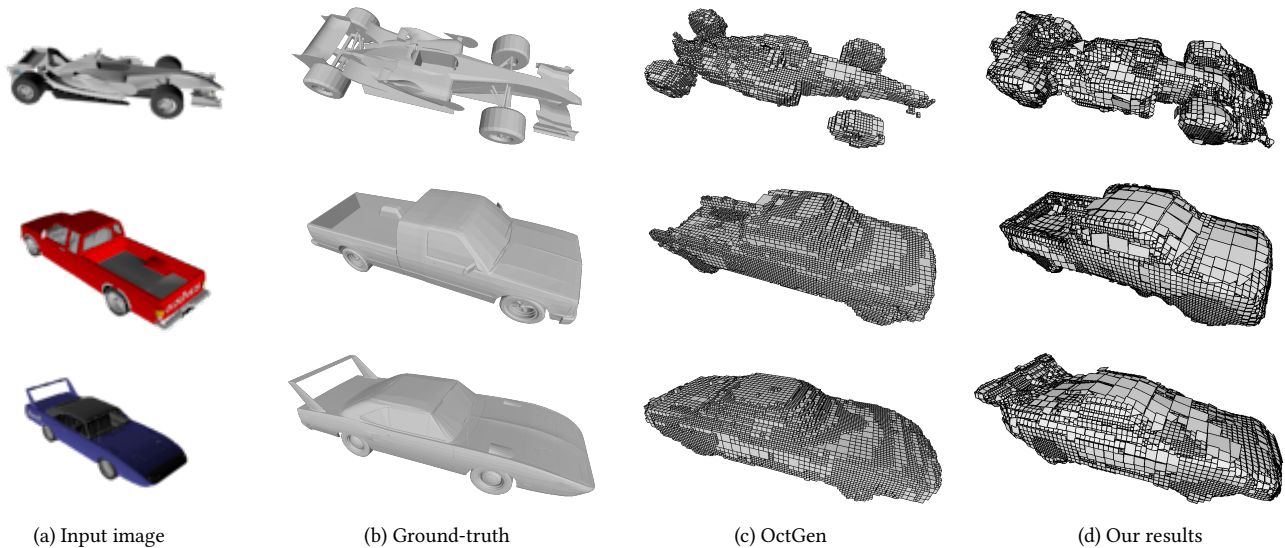

For computing the Chamfer distance for the output of OctGen,
we densely sample the points from the boundary octant boxes for evaluation.
Our Adaptive O-CNN has the lower Chamfer distance error than OctGen: 0.274 vs. 0.294.
A visual comparison is shown in \Cref{fig:image2shape-ogn}: our results preserve more details than OctGen, and
the resulting surface patches are much more faithful to the ground-truth, especially on the flat regions.

\begin{table*}[t]
  \centering
  \resizebox{0.96\textwidth}{!}{%
    \begin{tabular}{l|l|lllllllllllll}
      \toprule
      Method         & mean          & pla.          & ben.          & cab.          & car           & cha.          & mon.          & lam.          & spe.          & fir.          & cou.          & tab.          & cel.          & wat.          \\ \midrule
      PSG            & 6.41          & 3.36          & 4.31          & 8.51          & 8.63          & 6.35          & 6.47          & 7.66          & 15.9          & \textbf{1.58} & 6.92          & \textbf{3.93} & 3.76          & 5.94          \\
      AtlasNet(25)   & 5.11          & 2.54          & 3.91          & 5.39          & 4.18          & 6.77          & 6.71          & \textbf{7.24} & \textbf{8.18} & 1.63          & 6.76          & 4.35          & 3.91          & \textbf{4.91} \\
      Adaptive O-CNN & \textbf{4.63} & \textbf{2.45} & \textbf{2.69} & \textbf{2.67} & \textbf{1.80} & \textbf{6.13} & \textbf{6.27} & 10.92         & 9.43          & 1.68          & \textbf{4.42} & 4.19          & \textbf{2.51} & 5.04          \\
      \bottomrule
    \end{tabular}
  }
  \caption{Experimental statistics of shape reconstruction from a single image. The Chamfer distance in this table is multiplied by 1000 for better display. The mean is the category-wise average. Both the PSG and AtlasNet results are those provided in ~\cite{Groueix2018}. }
  \label{table:image2shape:chamfer} 
\end{table*}

%% file: src2/conclusion.tex
\section{Conclusion} \label{sec:conclusion}
We present a novel Adaptive O-CNN for 3D encoding and decoding. The encoder and decoder of Adaptive O-CNN utilize the nice properties of the patch-guided adaptive octree structure: compactness, adaptiveness, and high-quality approximation of the shape. We show the high memory and computational efficiency of Adaptive O-CNN, and demonstrate its superiority over other state-of-the-art methods including existing octree-based CNNs on some typical 3D learning tasks, including 3D autoencoding, surface completion from noisy and incomplete point clouds, and surface prediction from images. \looseness=-1

One limitation in our implementation is that the adjacent patches in the adaptive octree are not seamless. To obtain a well-connected mesh output, we need to use other mesh repairing or surface reconstruction techniques. In fact, we observe that most of the seams can be stitched by snapping the nearby vertices of adjacent patches. We would like to add a regularization loss function to reduce the seam, and develop a post-processing method to stitch all the gaps.

\rev{Another limitation is that the planar patch we used in Adaptive O-CNN does not approximate curved features very well, for instance, see the car wheel in \Cref{fig:ablation}.}
In the future, we would like to explore the use of non-planar surface patches in Adaptive O-CNN. Quadratic surface patches or its subclasses --- parabolic surface patches and ellipsoidal patches are promising patches because they have simple expressions and planes are a special case of them. Another direction is to use other fitting quality metrics to guide the subdivision of octants, for instance, using the topological similarity between the local fitted patch and the ground-truth surface patch as guidance to ensure that the fitted patch approximates the local shape well both in geometry and topology.

%% file: src2/appendix.tex
\appendix
\section*{Appendix: Parameter Setting of Adaptive O-CNN}

\begin{figure}[t]
  \centerline{
    \includegraphics[width=1\linewidth]{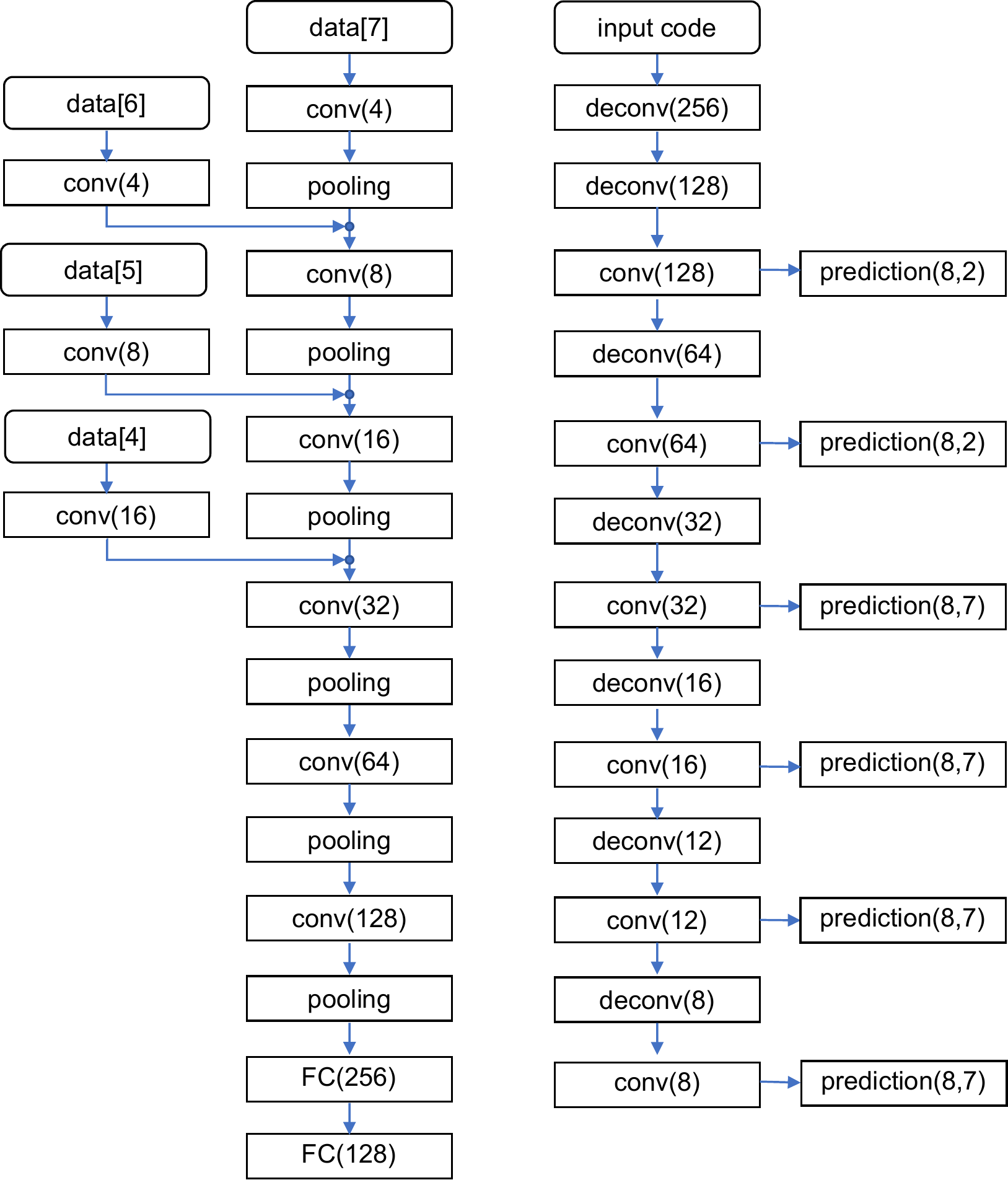}
  }
  \caption{Adaptive O-CNN encoder (left) and Adaptive O-CNN decoder (right) for an octree with a max-depth 7.}
  \label{fig:network} \vspace{-5mm}
\end{figure}

The detailed Adaptive O-CNN encoder and decoder networks for an octree with max-depth 7 is shown in ~\Cref{fig:network}.
In the figure,  $\mathrm{data}[d]$ represents the input feature at the $d^{th}$ octree level.
$\mathrm{conv}(c)$ represents the convolution operation with kernel size $3$ and output channel number $c$.
$\mathrm{deconv}(c)$ represents the deconvolution operation with kernel size $3$, stride $2$ and output channel number $c$.
The kernel size and stride of the $\mathrm{pooling}$ operation are both $2$.
$\mathrm{FC}(c)$ represents the fully connected layer with output channel number $c$.
$\mathrm{prediction}(c_1, c_2)$ is the \emph{prediction module} introduced in Section \textcolor[rgb]{1.0,0.0,0.0}{4.2}, which includes two $\mathrm{FC}(c)$ operations.
Here $c_1$ is the number of output channels of the first $\mathrm{FC(c)}$ operation, and is fixed to $8$.
And $c_2$ is the number of output channels of the second $\mathrm{FC(c)}$ operation, with which the plane parameters and the octant statuses are predicted.
Since we make the octree adaptive from the $4^{th}$ level, the value of $c_2$ at the second and the third level in ~\Cref{fig:network} is set to $2$, predicting whether to split an octant or not.
From the $4^{th}$ octree level the value of $c_2$ is set to $7$: $3$ channels of the output are used to predict the octant fitting status: \emph{empty}, \emph{surface-well-approximated}, and \emph{surface-poorly-approximated}; the other $4$ channels are used to regress the plane parameters $(\mn, d^\star)$.
The input latent code dimension of the decoder is set to 128.

We use the SGD solver to optimize the neural network, and the batch size is set to 32.
In the shape classification experiment, the initial learning rate is 0.1, and it is decreased by a factor of 10 after every 10 epochs, and stops after 40 epochs.
In the 3D autoencoding experiment, the initial learning rate is 0.1, and it is decreased by a factor of 10 after 100k, 200k, 250k iterations respectively, and stops after 350k iterations.
In the shape prediction from a single image task, the initial learning rate is 0.1, and it is decreased by a factor of 10 after 150k, 300k, 350k iterations respectively, and stops after 400k iterations.

